\def\year{2020}\relax
\documentclass[letterpaper]{article} 
\usepackage{aaai20-AIoT} 
\usepackage{times} 
\usepackage{helvet} 
\usepackage{courier} 
\usepackage[hyphens]{url} 
\usepackage{graphicx} 
 
\usepackage{amsmath}

\usepackage{algorithm,algpseudocode}
\usepackage{amsmath}
\usepackage{amsfonts}
\usepackage{amssymb}

\usepackage{mathtools}

\usepackage{graphicx} 
\title{Pairwise Neural Networks (PairNets) with Low Memory\\ for Fast On-Device Applications}
\author{
\\ \Large \textbf{Luna M. Zhang}\\  
}
\begin{document}

\maketitle

\begin{abstract}

A traditional artificial neural network (ANN) is normally trained slowly by a gradient descent algorithm, such as the backpropagation algorithm, since a large number of hyperparameters of the ANN need to be fine-tuned with many training epochs. Since a large number of hyperparameters of a deep neural network, such as a convolutional neural network, occupy much memory, a memory-inefficient deep learning model is not ideal for real-time Internet of Things (IoT) applications on various devices, such as mobile phones. Thus, it is necessary to develop fast and memory-efficient Artificial Intelligence of Things (AIoT) systems for real-time on-device applications. We created a novel wide and shallow 4-layer ANN called ``Pairwise Neural Network" (``PairNet") with high-speed non-gradient-descent hyperparameter optimization. The PairNet is trained quickly with only one epoch since its hyperparameters are directly optimized one-time via simply solving a system of linear equations by using the multivariate least squares fitting method.  In addition, an $n$-input space is partitioned into many $n$-input data subspaces, and a local PairNet is built in a local $n$-input subspace. This divide-and-conquer approach can train the local PairNet using specific local features to improve model performance. Simulation results indicate that the three PairNets with incremental learning have smaller average prediction mean squared errors, and achieve much higher speeds than traditional ANNs. An important future work is to develop better and faster non-gradient-descent hyperparameter optimization algorithms to generate effective, fast, and memory-efficient PairNets with incremental learning on optimal subspaces for real-time AIoT on-device applications. 
\end{abstract}

\section{Introduction}

Research in Artificial Neural Networks (ANNs) has had various important breakthroughs since the first work in ANNs was done in 1943 (McCulloch and Pitts 1943). In general, three major types of ANNs include the neuroscience-based ANN, the non-neuroscience-based ANN, and the hybrid ANN based on both neuroscience and other sciences. Brief overviews about the three ANNs are introduced as follows. 

The first important research problem is how to develop an effective ANN based on neuroscience and cognitive science. Hebb reinforced the artificial neurons defined by McCulloch and Pitts and showed how they worked in 1949 (Hebb 1949). It noted that neural pathways were strengthened each time that they were used. If two nerves fire simultaneously, then the connection between them becomes enhanced. The advanced Hebbian-LMS learning algorithm was developed in 2015 (Widrow et al. 2015).

The second important research problem is how to develop an effective ANN based on sciences other than neuroscience. In 1957, Rosenblatt invented the perceptron (Rosenblatt 1958). Unfortunately, the simple single-layer perceptron had limited ability for pattern recognition (Minsky and Papert, 1969).
In 1959, Widrow and Hoff developed new models called ADALINE and MADALINE. MADALINE (Many ADALINE) was the first neural network to be applied to real world problems (Widrow and Hoff 1992). In 1960, Widrow and Hoff developed the least mean squares (LMS) algorithm (Widrow et al. 1960). In early 1970s, Werbos developed the non-neuroscience-based backpropagation algorithm for training multilayer neural networks (Werbos 1974). Backpropagation is an efficient and precise technique in calculating all of the derivatives of a target quantity, such as pattern classification error with respect to a large set of input quantities, which may be weights in a neural network. The weights get optimized to minimize the loss function (Werbos 1990). Rumelhart, Hinton, and Williams publicized and described the backpropagation method for multilayer neural networks in 1986 (Rumelhart et al. 1986). In recent years, Deep Neural Networks (DNNs) with more hidden layers than shallow neural networks have many applications in computer vision (Larochelle et al. 2009; He et al. 2016; Szegedy et al. 2015), image processing (Krizhevsky et al. 2012), pattern recognition (Szegedy et al. 2017), bioinformatics (Esteva et al. 2017), etc. Deep learning is an important research area in machine learning and artificial intelligence that allows computational models with many processing layers to more accurately learn and model high-level abstractions from data (LeCun et al. 2015). An application is DeepMind's AlphaGo, a computer program that is very powerful in the game of Go (Silver et al. 2016). It uses neural networks as one of its techniques, with extensive training. Deep belief networks are a specific type of DNN that are probabilistic models with layers typically made of restricted Boltzmann machines (Hinton et al. 2006). In particular, DNNs, such as Convolutional Neural Networks (CNNs), typically take a very long time to be trained well. 

The third important research problem is how to develop an effective hybrid ANN based on both neuroscience and other sciences. 
For example, a new plastic neural network has a hybrid architecture based on properties of biological neural networks and a traditional ANN (Miconi et al. 2018). However, it still applies the slow backpropagation training algorithm to optimize weights of the plastic neural network.

The ANN and the hybrid ANN have hyperparameters to be optimized, such as weights between neurons, numbers of different layers, numbers of neurons on different layers, and different activation functions mapping summations of weighted inputs to outputs. An important research goal is to develop a new ANN with high computation speed and high performance, such as low validation errors, for various real-time machine learning applications. Some problems are discussed as follows.

First, backpropagation is a popular gradient descent-based training algorithm that is used to optimize weights, but it is a very slow optimization process which needs extensive training with many epochs. Other intelligent training algorithms use various advanced optimization methods, such as genetic algorithms (Loussaief and Abdelkrim 2018), and particle swarm optimization methods (Sinha et al. 2018) and to try to find optimal hyperparameters of an ANN. However, these commonly used training algorithms also require very long training times. 

Secondly, neural network structure optimization algorithms also take a lot of time to find optimal or near-optimal numbers of different layers and numbers of neurons on different layers. Especially, DNNs need much longer time. Thus, it is useful to develop fast wide and shallow neural networks with relatively small numbers of neurons on different layers for real-time machine learning applications. 

Traditionally, an ANN is trained very slowly by a gradient descent algorithm such as the backpropagation algorithm since a large number of hyperparameters of the ANN need to be fine-tuned with many training epochs. Therefore, the ANN's hyperparameter optimization challenge is how to develop high-speed non-gradient-descent training algorithms to optimize ANN's architecture. Many current DNNs, such as CNNs, occupy a lot of memory. Thus, it is necessary to develop fast and memory-efficient machine learning systems for real-time AIoT applications. 

For these long-term research problems related to building fast and memory-efficient systems of AIoT, we created a novel shallow 4-layer ANN called the Pairwise Neural Network (PairNet) (Zhang, 2019). In this paper, we created a new high-speed non-gradient-descent hyperparameter optimization algorithm with incremental learning for a PairNet with low memory for real-time AIoT applications.

\section{Pairwise Neural Network (PairNet)}

For a regression problem, the PairNet consists of four layers of neurons that map $n$ inputs on the first layer to one numerical output on the fourth layer. 

{\em Layer 1}: Layer 1 has $n$ neuron pairs to map $n$ inputs to $2n$ outputs. Each pair has two neurons where one neuron has an increasing activation function $g_{i}(x_{i})\in[0,1]$ that generates a positive normalized value, and the other neuron has a decreasing activation function $(1-g_{i}(x_{i}))$ that generates a negative normalized value for $i=1,2,...,n$. 

{\em Layer 2}: Layer 2 consists of $2^n$ neurons, where each neuron has an activation function to map $n$ inputs to an output as a complementary decision fusion. Each of the $n$ inputs is an output of one of the two neurons of each neuron pair on Layer 1. Let $g_{i}$ denote $g_{i}(x_{i})$, and $\bar{g}_{i}$ denote $(1-g_{i}(x_{i}))$ for $i=1,2,...,n$. Sample activation functions of neurons on Layer 2 are given as follows:
{
\begin{eqnarray}
w_{1}
=\alpha_{1}g_1+\alpha_{2}g_2+...+\alpha_{n-1}g_{n-1}+\alpha_{n}g_n,\nonumber
\\
w_{2}=
\alpha_{1}g_1+\alpha_{2}g_2+...+\alpha_{n-1}g_{n-1}+\alpha_{n}\bar{g}_{n},\nonumber
\\
......,\nonumber
\\ 
w_{2^{n}-1}=\alpha_{1}\bar{g}_1+\alpha_{2}\bar{g}_2+...+\alpha_{n-1}\bar{g}_{n-1}+\alpha_{n}g_n ,\nonumber
\\
w_{2^{n}}
=\alpha_{1}\bar{g}_1+\alpha_{2}\bar{g}_2+...+\alpha_{n-1}\bar{g}_{n-1}+\alpha_{n}\bar{g}_n,\
\end{eqnarray}
}where $\alpha_{i}$ are hyperparameters to be optimized for $0\leq \alpha_{i}\leq 1$, $i=1,2,...,n$, and $\sum_{i=1}^{n}\alpha_{i}=1$. $\sum_{k=1}^{2^{n}}w_k=2^{n-1}\sum_{i=1}^{n}\alpha_{i}(g_i+\bar{g}_i)
=2^{n-1}\sum_{i=1}^{n}\alpha_{i}=2^{n-1}$. For a special case, equal weights $\alpha_{i}=\frac{1}{n}$ for $i=1,2,...,n$.

{\em Layer 3}: Layer 3 also consists of $2^n$ neurons but transforms the outputs of the second layer to $2^n$ individual output decisions. 
\begin{eqnarray}
w_k=
1+\displaystyle \frac{y_{k}^{1}-c_{k}}{\eta_{k}}
&\mbox{for\hspace{0.1in}$ (c_{k}-\eta_{k})
\leq y_{k}\leq c_{k}$},
\end{eqnarray}
\begin{eqnarray}
w_k=1-\displaystyle \frac{y_{k}^{2}-c_{k}}{\delta_{k}}
&\mbox{for\hspace{0.1in}$ c_{k}
\leq y_{k}\leq (c_{k}+\delta_{k})$},
\end{eqnarray}
where $k=1,2,...,2^{n}$. $\bar{y}_k$, sample activation functions of  neurons on Layer 3, are defined as
\begin{eqnarray}
\bar{y}_k = \frac{y_{k}^{1} + y_{k}^{2}}{2} = c_{k} + \frac{(1 - w_k)\gamma_k}{2},
\end{eqnarray}
where $\gamma_k = \delta_k - \eta_k$. 

{\em Layer 4}: Layer 4 calculates a final output decision by computing a weighted average of the $2^n$ individual output decisions of Layer 3. $f(x_1,x_1,...,x_n)$, a sample activation function of  the output neuron on Layer 4, is given by
\begin{eqnarray}
f(x_1,x_1,...,x_n)=\sum_{k=1}^{2^{n}}
\beta_k \bar{y}_k,\
\end{eqnarray}
where $\beta_k = \frac{w_k}{\sum_{j=1}^{2^{n}}w_j}=\frac{w_k}{2^{n-1}}$.

For convenience, we have
\begin{eqnarray}
f(x_1,x_1,...,x_n)=\nonumber \\ 
\bar{f}(x_1,x_1,...,x_n)+\tilde{f}(x_1,x_1,...,x_n),\
\end{eqnarray}
where 
\begin{eqnarray}
\bar{f}(x_1,x_1,...,x_n)=\sum_{k=1}^{2^{n}}\beta_{k}c_k,\
\end{eqnarray}
\begin{eqnarray} 
\tilde{f}(x_1,x_1,...,x_n)=\sum_{k=1}^{2^{n}}\beta_{k}\theta_{k}\gamma_k,
\end{eqnarray}
where $\theta_{k}=\frac{1 - w_{k}}{2}$ for $k=1,2,...,2^{n}$. 

We have
\begin{eqnarray}
f(x_1,x_1,...,x_n)= \sum_{k=1}^{2^{n}}(\beta_{k}c_k + \beta_{k}\theta_{k}\gamma_k).\
\end{eqnarray}

Finally, the PairNet $f(x_1,x_1,...,x_n)$ consists of the linear $\bar{f}( x_1,x_1,...,x_n)$
and the nonlinear $\tilde{f}(x_1,x_1,...,x_n)$. 

A 3-input-1-output PairNet is shown in Fig. 1.

\begin{figure}[h]
\centering
\includegraphics[width=0.9\columnwidth]{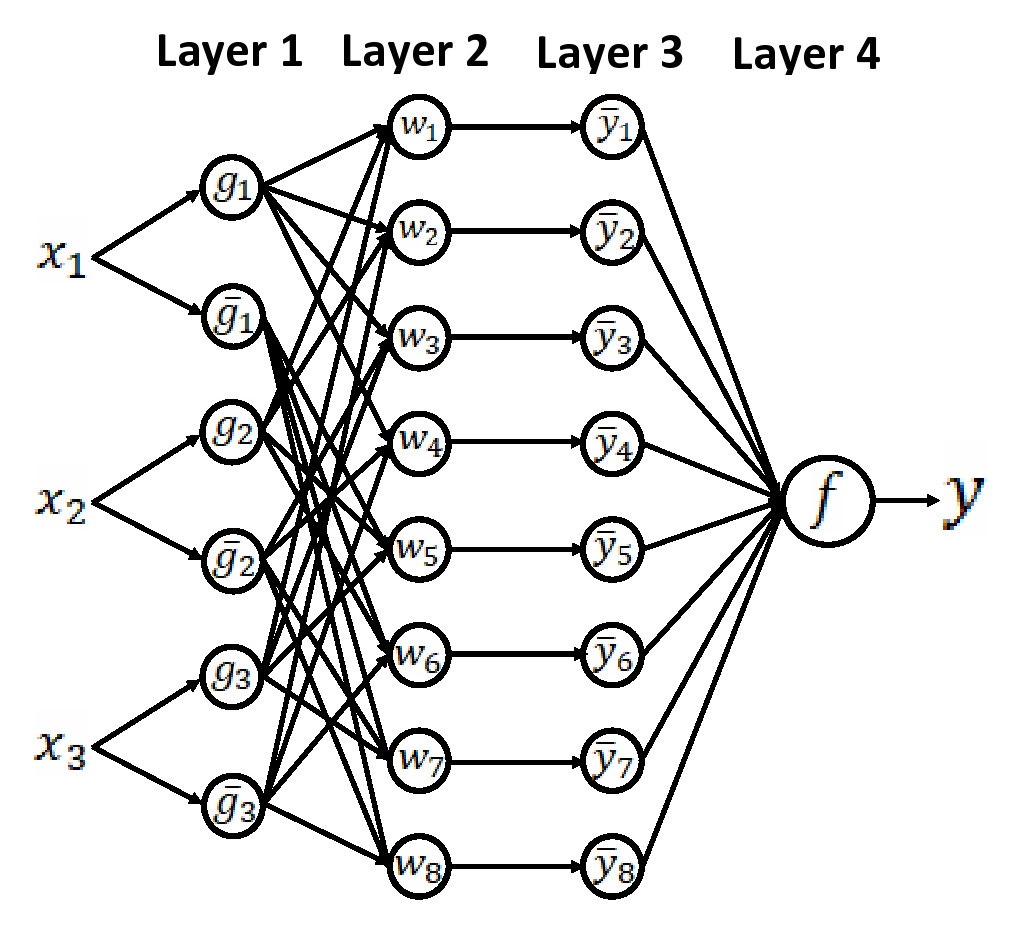} 
\caption{A 3-input-1-output PairNet}
\label{fig1}
\end{figure}

\section{Fast Training Algorithm with Hyperparameter Optimization}

We develop a new fast multivariate least-squares algorithm to directly find optimal hyperparameters for the best-fitting model by quickly solving a system of linear equations for a given training dataset. For $n$ inputs, $2^{n+1}$ linear equations need to be solved to get $2^{n+1}$ hyperparameters to minimize the mean squared error (MSE). Significantly, gradient descent training with a large number of epochs is not needed at all. The PairNet is quickly trained with only one epoch using the multivariate least squares fitting method since more epochs are not applicable.

A data set has $n$ inputs $x_i$ for $i=1,2,...,n$, and one output $y$. It has
$N$ data. An input $x_i$ has $m_i$ intervals in $[a_i, b_i]$ such that 
$[a_i, a_{i1}]$, $[a_{i1}, a_{i2}]$, ..., $[a_{im_{i}-2}, a_{im_{i}-1}]$, and $[a_{im_{i}}, b_i]$ for $m_i\geq1$, and $i=1,2,...,n$. Then there are $M$
($M=\prod_{i=1}^{n}m_i$) $n$-dimensional subspaces $S_{j}$ for $j=1,2,...,M$. $N$ data are distributed in the $M$ $n$-dimensional subspaces. A $n$-dimensional subspace $S_{j}$ has $N_{j}$ data with $N_{j}$ outputs $Y^j_p$ for $j=1,2,...,M$, $p=1,2,...,N_{j}$, and $N=\sum_{j=1}^{M}N_{j}$. 
For each $n$-dimensional subspace such as ($[a_11, a_{11}]$, $[a_{21}, a_{22}]$, ..., $[a_{n-11}, a_{n-12}]$, and $[a_{n1}, a_{n2}]$), 
a PairNet can map $n$ inputs $x_i$ for $i=1,2,...,n$ to one output $f_{j}(x_{1},...,x_{n})$ for $j=1,2,...,M$. Thus, an $n$-input space is partitioned into many $n$-input data subspaces, and a local PairNet is built in a local $n$-input subspace. This divide-and-conquer approach can train the local PairNet using specific local features to improve model performance. 

The objective optimization function for a PairNet $f_{j}(x_{1},...,x_{n}$ for $j=1,2,...,M$ is given below:
\begin{equation}
Q=\frac{1}{2}\sum_{p=1}^{N_{j}}[Y^j_p-f_{j}(x_{1_p},x_{2_p},...,x_{n_p})]^2.\
\end{equation}
\begin{equation}
Q=\frac{1}{2}\sum_{p=1}^{N_{j}}[Y^j_p-\sum_{k=1}^{2^{n}}(\beta^j_{k_p}c^j_{k} + \beta^j_{k_p}\theta^j_{k_p}\gamma^j_{k})]^2.\
\end{equation}

To minimize $Q$ by optimizing $2^{n+1}$ parameters ($c^j_k$ and $\gamma^j_k$) for $k=1,2,...,2^{n}$, we have
\begin{eqnarray}
\left\{\begin{array}{lll}
\frac{\partial Q}{\partial c^j_k}=0\\
\frac{\partial Q}{\partial \gamma^j_k}=0
,\\
\end{array}
\right.
\end{eqnarray}

then we have
{\footnotesize
\begin{eqnarray}
\left\{\begin{array}{lll}
\sum_{p=1}^{N}\beta^j_{k_p}(Y^j_p - \sum_{q=1}^{2^{n}}(\beta^j_{q_p}c_q + \beta^j_{q_p}\theta^j_{q_p}\gamma^j_q))
=0\\
\sum_{p=1}^{N}\beta^j_{k_p}\theta^j_{k_p}(Y^j_p - \sum_{q=1}^{2^{n}}(\beta^j_{q_p}c_q + \beta^j_{q_p}\theta^j_{q_p}\gamma^j_q))
=0
.\\
\end{array}
\right.
\end{eqnarray}
}

We have $2^{n+1}$ linear equations with $2^{n+1}$ hyperparameters ($c_k$ and $\gamma_k$) for $k=1,2,...,2^{n}$ as follows:

{\scriptsize
\begin{eqnarray}
\left\{\begin{array}{lll}
\sum_{p=1}^{N} \beta^j_{1_p}\sum_{q=1}^{2^{n}}(\beta^j_{q_p}(c^j_q + \theta^j_{q_p}\gamma^j_q))
=\sum_{p=1}^{N}\beta^j_{1_p}Y^j_p. \\
\sum_{p=1}^{N} \beta^j_{2_p}\sum_{q=1}^{2^{n}}(\beta^j_{q_p}(c^j_q + \theta^j_{q_p}\gamma^j_q))
=\sum_{p=1}^{N}\beta^j_{2_p}Y^j_p. \\

...\\
\sum_{p=1}^{N} \beta^j_{2^n_p}\sum_{q=1}^{2^{n}}(\beta^j_{q_p}(c^j_q + \theta^j_{q_p}\gamma^j_q))
=\sum_{p=1}^{N}\beta^j_{2^n_p}Y^j_p. \\
\sum_{p=1}^{N} \beta^j_{1_p}\theta^j_{1_p}\sum_{q=1}^{2^{n}}(\beta^j_{q_p}(c^j_q + \theta^j_{q_p}\gamma^j_q))
= \sum_{p=1}^{N}\beta^j_{1_p}\theta^j_{1_p}Y^j_p.\\
\sum_{p=1}^{N} \beta^j_{2_p}\theta^j_{2_p}\sum_{q=1}^{2^{n}}(\beta^j_{q_p}(c^j_q + \theta^j_{q_p}\gamma^j_q))
= \sum_{p=1}^{N}\beta^j_{2_p}\theta^j_{2_p}Y^j_p.\\
...\\
\sum_{p=1}^{N} \beta^j_{2^n_p}\theta^j_{2^n_p}\sum_{q=1}^{2^{n}}(\beta^j_{q_p}(c^j_q + \theta^j_{q_p}\gamma^j_q))
= \sum_{p=1}^{N}\beta^j_{2^n_p}\theta^j_{2^n_p}Y^j_p.\\
\end{array}
\right.
\end{eqnarray}
}

The fast hyperparameter optimization algorithm for creating $M$ PairNet models in $M$ subspaces is given in Algorithm 1.\\

\begin{algorithm}[H] 
\caption{Fast Hyperparameter Optimization Algorithm for Generating PairNet Models in Different Subspaces }
\label{alg:loop}
\begin{algorithmic}[1]
\Require{$m_i$ (the number of intervals of each input $x_i$) for $i=1,2,...,n$} 
\Ensure{optimized hyperparameters $c^j_l$ and
$\gamma^j_l$ for $l=1,2,...,2^{n}$ for a subspace $S_j$ for $j=1,2,...,M$}

\Statex

\For{$j = 1$ to $M$} 
\State {For each subspace $S_j$, calculate hyperparameters $c^j_l$ and
$\gamma^j_l$ for $l=1,2,...,2^{n}$ based on Eq. (14).}
\EndFor
\State \Return {$M$ PairNet Models $f_{j}(x_1,x_1,...,x_n)$ for $j=1, 2,..., M$.}

\end{algorithmic}
\end{algorithm}

A simple PairNet model selection algorithm with a random search method is given in Algorithm 2. 

\begin{algorithm}[H] 
\caption{ PairNet Model Selection Algorithm with Random Search}
\label{alg:loop}
\begin{algorithmic}[1]
\Require{$K$: the number of candidate PairNet models} 
\Ensure{the best PairNet model} 

\Statex
\State {Randomly generate $M$ subspaces $S_j$ 
for $j=1,2,...,M$. }
\State {Run \bf{Algorithm 1}.}
\State {Evaluate the performance of the newly generated PairNet model with $M$ local PairNet models for 
the $M$ subspaces.}
\State {Set the best PairNet model as the newly generated PairNet model.}

\For{$k = 1$ to $K$} 
\State {Randomly generate $M_k$ subspaces $S^k_j$ for $j=1,2,...,M_k $. }
\State {Run \bf{Algorithm 1}.}
\State {Evaluate the performance of the newly generated PairNet model.}
\State {If the newly generated PairNet model is better than the best PairNet model, then the best PairNet model is the newly generated PairNet model.}
\EndFor
\State \Return {the best PairNet model.}

\end{algorithmic}
\end{algorithm}

Algorithm 3, a fast incremental learning method, can quickly train a local PairNet for new real-time training data. The saved optimized hyperparameters of Eq. (14) are re-used for future real-time incremental learning. If the number of inputs is not large, then  the optimized hyperparameters need small memory. Thus, the fast and memory-efficient PairNets with incremental learning are suitable for real-time on-device AIoT applications. 

\begin{algorithm}[H] 
\caption{ Incremental Learning Algorithm for PairNets}
\label{alg:loop}
\begin{algorithmic}[1]
\Require{$K$: the number of candidate PairNet models } 
\Ensure{the trained PairNets for real-time prediction} 

\Statex
\State {Run Algorithm 2 using $K$ to initially pre-train all local PairNets in all subspaces by using currently available training data.}
\For{each new training data $d$} 
\State {Find the appropriate subspace for $d$. }
\State {Use $d$ to solve Eq. (14) to update the hyperparameters of a local PairNet in the subspace, and save them.}
\State {Create a new local PairNet with the optimized hyperparameters in the subspace.}
\EndFor
\State \Return {trained local PairNets for real-time prediction.}

\end{algorithmic}
\end{algorithm}

Algorithm 4 shows the basic steps of using the trained PairNet for real-time prediction. 

\begin{algorithm}[H] 
\caption{Real-time PairNet Prediction Algorithm}
\label{alg:loop}
\begin{algorithmic}[1]
\Require{real-time input data $d$} 
\Ensure{a predicted value} 

\Statex
\State {Find an appropriate subspace for $d$.}
\State {Use the trained local PairNet in the appropriate subspace to map $d$ to a predicted value. }
\State \Return {a predicted value.}

\end{algorithmic}
\end{algorithm}

\section{Performance Analysis for Real-time Time Series Prediction}

Daily time series data starting on 7/1/1954 (Historical Data and Trend Chart of Effective Federal Funds Rate) were converted into 16,185 3-input-1-output training data, 50 testing data, 75 testing data, and 100 testing data. Incremental learning was done only on the testing dataset. Each daily test data became the new training data $d$ for Algorithm 3 to evaluate its performance. The inputs are 3 consecutive days' rates, and the output is the 4th day's rate. The range of the training data inputs is [0.13, 22.36]. A traditional ANN with incremental learning is noted as $ANN_{IL}$. The three-input-one-output PairNet is denoted as $PairNet_{ijk}$ with $i\times{j}\times{k}$ subspaces, where the three inputs has $i$ intervals, $j$ intervals, and $k$ intervals.
Initially, an $ANN_{IL}$ was pre-trained by using 16,185 training data. The $ANN_{IL}$ has two hidden layers with 50 neurons on each layer. Algorithm 3 using $K$=200 was used to pre-train PairNets with 2, 4 and 8 subspaces by using 16,185 training data. Then, the pre-trained $ANN_{IL}$ and the pre-trained PairNet performed daily incremental learning; they were trained by using a new daily data. Finally, the incrementally-trained $ANN_{IL}$ predicted the next day's rate, and Algorithm 4 was used for an appropriate local PairNet to predict the next day's rate. Two even intervals ([0.13, 11.245] and [11.245, 22.36]) are used.

Table 1 shows the average prediction MSEs of $ANN_{IL}$ using different training epochs, and average prediction MSEs of seven PairNets with different subspaces for $N$ testing data. Simulation results shown in Table 1 indicate that the seven PairNets have smaller average prediction MSEs than the six $ANN_{IL}$ models. $PairNet_{222}$ with 8 subspaces achieves all three lowest prediction MSEs. 

The overall average prediction MSEs of PairNets with 2 subspaces, 4 subspaces, and 8 subspaces are 0.0612, 0.0582, and 0.0536, respectively. For this case, the more subspaces a PairNet has, the more accurate it makes predictions. More research on the relationship between the nunber of subspaces and performance of a PairNet will be done with more simulations. The overall average prediction MSE of the six $ANN_{IL}$ models is 0.0743. Thus, the PairNets perform better than the $ANN_{IL}$ models based on the overall average prediction MSEs.

\begin{table}[h!]
\caption{Average Prediction MSEs of $ANN_{IL}$ and PairNets}\smallskip
\centering
\resizebox{1.0\columnwidth}{!}{
\smallskip\begin{tabular}{|l|l|l|l|l|}
Neural Network  & Epochs & $N$ =  50 & $N$ =  75  & $N$ =  100 \\
$ANN_{IL}$&100 & 0.0708 & 0.0828 & 0.0727 \\
$ANN_{IL}$&200 & 0.0598 & 0.0909 & 0.0776\\
$ANN_{IL}$&300 & 0.0636 & 0.0813 & 0.0694\\
$ANN_{IL}$&1000 & 0.0609 & 0.0827 & 0.0746 \\
$ANN_{IL}$&2000 & 0.0703 & 0.0857 & 0.0847\\
$ANN_{IL}$&3000 & 0.0621 & 0.0808 & 0.0675\\
$PairNet_{112}$&1 & 0.0536& 0.0727 & 0.0617\\
$PairNet_{121}$&1 & 0.0458& 0.0679 & 0.0579\\
$PairNet_{211}$&1 & 0.0547& 0.0728 & 0.0636\\
$PairNet_{122}$&1 & 0.0478& 0.0693 & 0.0587\\
$PairNet_{212}$&1 & 0.0502& 0.0670 & 0.0579\\
$PairNet_{221}$&1 & 0.0465& 0.0677 & 0.0588\\
$PairNet_{222}$&1 & 0.0448& 0.0624 & 0.0535\\
\end{tabular}
}
\label{table1}
\end{table}

Table 2 shows the average daily training times of the six $ANN_{IL}$ models, and those of the seven PairNets. 
Simulation results indicate that the seven PairNets achieve much higher speeds than the six $ANN_{IL}$ models.

\begin{table}[h!]
\caption{Average Daily Training Times (seconds)}\smallskip
\centering
\resizebox{1.0\columnwidth}{!}{
\smallskip\begin{tabular}{|l|l|l|l|l|}
Neural Network  & Epochs & $N$ =  50 & $N$ =  75  & $N$ =  100 \\
$ANN_{IL}$&100 & 0.15488 & 0.17605 & 0.17190 \\
$ANN_{IL}$&200 & 0.34944 & 0.34096 & 0.35673\\
$ANN_{IL}$&300 & 0.59982 & 0.56120 & 0.57743\\
$ANN_{IL}$&1000 & 1.97139 & 1.99363 & 2.04981 \\
$ANN_{IL}$&2000 & 4.05327 & 4.44860 & 3.97650\\
$ANN_{IL}$&3000 & 6.83142 & 6.71835 & 6.38905\\
$PairNet_{112}$&1 & 0.00083& 0.00080 & 0.00065\\
$PairNet_{121}$&1 & 0.00099& 0.00097 & 0.00104\\
$PairNet_{211}$&1 & 0.00093& 0.00104 & 0.00064\\
$PairNet_{122}$&1 & 0.00130& 0.00062 & 0.00106\\
$PairNet_{212}$&1 & 0.00129& 0.00086 & 0.00096\\
$PairNet_{221}$&1 & 0.00107 & 0.00033 & 0.00114\\
$PairNet_{222}$&1 & 0.00095& 0.00124 & 0.00056\\
\end{tabular}
}
\label{table1}
\end{table}
 
A PairNet with 2 subspaces ($PairNet_{112}$, $PairNet_{121}$, and $PairNet_{211}$), denoted as $PairNet^{1}$, needs 14KB memory for its hyperparameters. A PairNet with 4 subspaces ($PairNet_{122}$, $PairNet_{212}$, and $PairNet_{221}$), denoted as $PairNet^{2}$, needs 28KB memory for its hyperparameters. $PairNet_{222}$ needs 42KB memory for its hyperparameters. The code for a PairNet needs 19KB memory. The memory sizes of trained $ANN_{IL}$ models with different numbers of hidden layers (50 neurons on each layer) are given in Table 3. Thus, the 3-hidden-layer PairNet is much more memory-efficient than the $ANN_{IL}$ models. The $ANN_{IL}$ with 50 hidden layers needs 1,867KB memory. DNNs with more than 50 hidden layers will need much bigger memory. Thus, the fast and memory-efficient PairNet is more suitable than a traditional ANN for various devices with small memory that are used in real-time AIoT on-device applications.  

\begin{table}[h!]
\caption{Memory Sizes of the $ANN_{IL}$ and the PairNets}\smallskip
\centering
\resizebox{1.0\columnwidth}{!}{
\smallskip\begin{tabular}{|l|l|l|l|l|}
Neural Network  &Hidden Layers & Memory (KB)  \\
$PairNet^{1}$& 3 & 33\\
$PairNet^{2}$& 3 & 47\\
$PairNet_{222}$& 3 & 61\\
$ANN_{IL}$&3 & 101 \\
$ANN_{IL}$&5 & 176 \\
$ANN_{IL}$&10 & 363 \\
$ANN_{IL}$&20 & 740 \\
$ANN_{IL}$&50 & 1867 \\
\end{tabular}
}
\label{table1}
\end{table}

\section{Conclusions}
Different from slow gradient descent training algorithms and other tedious training algorithms, such as genetic algorithms, the new high-speed non-gradient-descent training algorithm with direct hyperparameter computation can quickly train the new wide and shallow 4-layer PairNet with only one epoch since its hyperparameters are directly optimized one-time via simply solving a system of linear equations by using the multivariate least squares fitting method. For AIoT applications, partitioning big data space into many small data subspaces is useful and easy to build local PairNets because having nonlinear functions on small data subspaces are simpler than having a global function on the whole big data space. 

Simulation results indicate that the PairNets have smaller average prediction MSEs, and achieve much higher speeds than the ANNs for the real-time time series prediction application. Thus, it is feasible and necessary to continue to improve the effectiveness and efficiency of the new shallow PairNet by developing more intelligent non-gradient-descent training algorithms for real-time AIoT applications.

\section{Future Works}

The PairNet's performance will be compared with that of other existing techniques, such as LSTM-based recurrent neural network, and more datasets will be used.
Although the PairNet is a shallow neural network since it has only four layers of neurons, it is actually a wide neural network because both the second layer and the third layer have $2^n$ neurons with the first layer having $n$ neurons. Thus, the PairNet has the curse of dimensionality. We will develop advanced divide-and-conquer methods to solve the problem. 

The preliminary simulations applied an even data partitioning method to divide a whole $3$-input space into subspaces. In the future, more intelligent space partition methods will be created to build more effective local PairNets on optimized $n$-input data subspaces. Also, for classification problems, the new PairNet with a new activation function, such as Softmax, of the neuron on Layer 4 will be created. 

Furthermore, for image classification problems, the new PairNet (for classification problems) can replace the fully connected layer of a CNN with the goal of making a fast CNN with high performance. It will be evaluated by solving commonly used benchmark classification problems using datasets like MNIST, CIFAR10, and CIFAR100.

In summary, a significant future work is to develop better and faster non-gradient-descent hyperparameter optimization algorithms to generate effective, fast and memory-efficient PairNets on optimal subspaces for real-time AIoT on-device applications.    

\section{Acknowledgments}
The author would like to thank the reviewers very much for their valuable comments that help improve the quality of this paper. 

\section{References} 

\smallskip \noindent Esteva, A.; Kuprel, B.; Novoa, R.A.; Ko, J.; Swetter, S.M.; Blau, H.M.; and Thrun, S. 2017. Dermatologist-level classification of skin cancer with deep neural networks. \textit{Nature} 542(7639): 115--118.

\smallskip \noindent He, K.; Zhang, X.; Ren, S.; and Sun, J. 2016. Deep Residual Learning for Image Recognition. In \textit{Proceedings of  the 2016 IEEE Conference on Computer Vision and Pattern Recognition,} 770--778.

\smallskip \noindent Hebb, D.O. 1949. \textit{The Organization of Behavior: A Neuropsychological Theory.} New York: John Wiley \& Sons, Inc.

\smallskip \noindent Hinton, G.; Osindero, S.; and Teh, Y.-W. 2006. A fast learning algorithm for deep belief nets. \textit{Neural Computation} 18(7): 1527--1554.

\smallskip \noindent  Historical Data and Trend Chart of Effective Federal Funds Rate. $https://www.forecasts.org/data/data/DFF.htm$.

\smallskip \noindent Krizhevsky, A.; Sutskever, I.; and Hinton, G.E. 2012. Imagenet classification with deep convolutional neural networks. In \textit{Advances in Neural Information Processing Systems 25,} 1097--1105. 

\smallskip \noindent Larochelle, H.; Bengio, Y.; Louradour, J.; and Lamblin, P. 2009. Exploring Strategies for Training Deep Neural Networks. \textit{Journal of Machine Learning Research} 10: 1--40.

\smallskip \noindent LeCun, Y.; Bengio, Y.; and Hinton, G.E. 2015. Deep learning. \textit{Nature} 521: 436--444. 

\smallskip \noindent Loussaief, S.; and Abdelkrim, A. 2018. Convolutional Neural Network Hyper-Parameters Optimization based on Genetic Algorithms. \textit{International Journal of Advanced Computer Science and Applications} 9(10): 252--266.

\smallskip \noindent McCulloch, W. and Pitts, W. 1943. A logical calculus of the ideas immanent in nervous activity. \textit{Bulletin of Mathematical Biophysics} 5:115--133.

\smallskip \noindent Miconi, T.; Clune, J.; and Stanley, K. O. 2018. Differentiable plasticity: training plastic neural networks with backpropagation. In \textit{Proceedings of the 35th International Conference on Machine Learning,} arXiv:1804.02464.

\smallskip \noindent Minsky, M., and Papert, S. 1972. (2nd edition with corrections, first edition 1969) \textit{Perceptrons: An Introduction to Computational Geometry.} The MIT Press, Cambridge MA.

\smallskip \noindent Rosenblatt, F. 1958. The Perceptron: A Probabilistic Model for Information Storage and Organization in the Brain. \textit{Psychological Review} 65(6): 386--408.

\smallskip \noindent Rumelhart, D. E.; Hinton, G. E.; and Williams, R. J. 1986. Learning representations by back-propagating errors. \textit{Nature} 323: 533--536.

\smallskip \noindent Silver,  D.; Huang, A.; Maddison, C. J.; Guez, A.; Sifre, L.; Driessche, G. V. D.; Schrittwieser, J.; Antonoglou, I.; Panneershelvam, V.; Lanctot, M.; Dieleman, S.; Grewe, D.; Nham, J.; Kalchbrenner, N.; Sutskever, I.; Lillicrap, T.; Leach, M.; Kavukcuoglu, K.; Graepel, T.; and Hassabis, D. 2016. Mastering the game of Go with deep neural networks and tree search. \textit{Nature} 529: 484--503.

\smallskip \noindent Sinha, T.; Haidar, A.; and Verma, B. 2018. 
Particle Swarm Optimization Based Approach for Finding Optimal Values of Convolutional Neural Network Parameters. In \textit{Proceedings of 2018 IEEE Congress on Evolutionary Computation,} 1--6.

\smallskip \noindent Szegedy, C.; Ioffe, S.; Vanhoucke, V.; and Alemi, A. 2017. Inception-v4, Inception-ResNet and the Impact of Residual Connections on Learning. In \textit{Proceedings of the Thirty-First AAAI Conference on Artificial Intelligence (AAAI-17),} 4278--4284.

\smallskip \noindent Szegedy, C.; Liu, W.; Jia, Y.; Sermanet, P.; Reed S.; Anguelov D.; Erhan, D.; Vanhoucke, V.; and Rabinovich, A. 2015. Going Deeper with Convolutions. In \textit{Proceedings of 2015 IEEE Conference on Computer Vision and Pattern Recognition,} 1--9. 

\smallskip \noindent Werbos, P. 1990. Backpropagation through time: what it does and how to do it. \textit{Proceedings of the IEEE} 78(10): 1550--1160, .

\smallskip \noindent Werbos, P. 1974. Beyond Regression: New Tools for Prediction and Analysis in the Behavioral Sciences. PhD thesis, Harvard University. 

\smallskip \noindent Widrow B., and Hoff M.E. Jr. 1960. Adaptive switching circuits. In \textit{Proceedings of IRE WESCON Conf. Rec.,} part 4, 96--104.

\smallskip \noindent Widrow B.; Kim Y.; and Park D. 2015. The Hebbian-LMS Learning Algorithm. \textit{IEEE Computational Intelligence Magazine} 37--53.

\smallskip \noindent Widrow, B., and Lehr, M.A. 1993. Artificial Neural Networks of the Perceptron, Madaline, and Backpropagation Family. Proceedings of the Workshop on NeuroBionics, Goslar, Germany. \textit{Neurobionics} 133--205, H.-W. Bothe, M. Samii, R. Eckmiller, eds., North-Holland, Amsterdam. 

\smallskip \noindent Zhang L. M. 2019. PairNets: Novel Fast Shallow Artificial Neural Networks on Partitioned Subspaces. The 1st Workshop on Sets and Partitions at the 33rd Annual Conference on Neural Information Processing Systems (NeurIPS 2019), Vancouver, Dec. 14, 2019.  

\end{document}